\documentclass[11pt,a4paper]{article}
\usepackage[hyperref]{emnlp-ijcnlp-2019}
\usepackage{times}
\usepackage{latexsym}

\usepackage{amsfonts, bm} % formula
\usepackage{amsmath} % formula
\usepackage{float}
\usepackage{array}
\usepackage{booktabs} % tables
\usepackage[english]{babel}
\usepackage{multirow,graphicx,paralist}
\usepackage{makecell}
\usepackage{url}

\usepackage{framed}

\aclfinalcopy % Uncomment this line for the final submission

\newcommand{\model}{QAInfomax }
\newcommand{\clocal}{LC }
\newcommand{\cglobal}{GC }

\title{QAInfomax: Learning Robust Question Answering System \\by Mutual Information Maximization}

\author{Yi-Ting Yeh\quad Yun-Nung Chen \\
Department of Computer Science and Information Engineering\\
National Taiwan University, Taipei, Taiwan \\
\texttt{r07922064@csie.ntu.edu.tw\quad y.v.chen@ieee.org} %\\\And
\\}

\date{}

\begin{document}
\maketitle
\begin{abstract}

Standard accuracy metrics indicate that modern reading comprehension systems have achieved strong performance in many question answering datasets. 
However, the extent these systems truly understand language remains unknown, and existing systems are not good at distinguishing distractor sentences, which look related but do not actually answer the question.
To address this problem, we propose QAInfomax as a regularizer in reading comprehension systems by maximizing mutual information among passages, a question, and its answer. 
QAInfomax helps regularize the model to not simply learn the superficial correlation for answering questions.
The experiments show that our proposed QAInfomax achieves the state-of-the-art performance on the benchmark Adversarial-SQuAD dataset\footnote{The source code is publicly available at \url{https://github.com/MiuLab/QAInfomax}.}.

\end{abstract}

\section{Introduction}
\label{sec:intro}
Question answering tasks are widely used for training and testing machine comprehension and reasoning \cite{rajpurkar2016squad,joshi2017triviaqa}. 
However, high performance in standard automatic metrics has been achieved with only superficial understanding, as models exploit simple correlations in the data that happen to be predictive on most test examples.
\citet{jia2017adversarial} addressed this problem and proposed an adversarial version of the SQuAD dataset, which was created by adding a distractor sentence to each paragraph. 
The distractor sentences challenge the model robustness, and the created Adversarial-SQuAD data shows the inability of a model about distinguishing a sentence that actually answers the question from one that merely has words in common with it, where almost all state-of-the-art machine comprehension systems are significantly degraded on adversarial examples.

% 好像有點冗，但主要idea是這篇來的...
\citet{lewis2018generative} argued that over-fitting to superficial biases is partially caused by discriminative loss functions, which saturate when simple correlations allow the question to be answered confidently, leaving no incentive for further learning on the example.
%Therefore, they designed generative QA models to learn a prior over answers and a conditional language model for generating the question, and showed the improvement on Adversarial-SQuAD because the model tends to explain all question words, some of which may be unlikely under the distractor.
Therefore, they designed generative QA models, which use a generative loss function in question answering instead, and showed the improvement on Adversarial-SQuAD.
% 最後一句有點看不懂

Instead of regularizing models by generative loss functions, we propose an alternative approach named ``QAInfomax'' by maximizing mutual information (MI) among passages, questions, and answers, aiming at helping models be not stuck with superficial biases in the data during learning.
To efficiently estimate MI, QAInfomax incorporates the recently proposed deep infomax (DIM) in the model~\cite{hjelm2018learning}, which was proved effective in learning representations for image, audio \cite{ravanelli2018learning}, and graph domains \cite{velivckovic2018deep}.
In this work, the proposed QAInfomax further extends DIM to the text domain, and encourages the question answering model to generate answers carrying information that can explain not only questions but also itself, and thus be more sensitive to distractor sentences. 
Our contributions are summarized:
\begin{itemize}
    \item This paper first attempts at applying DIM-based MI estimation as a regularizer for representation learning in the NLP domain.
    \item The proposed QAInfomax achieves the state-of-the-art performance on the Adversarial-SQuAD dataset without additional training data, demonstrating its better robustness.
\end{itemize}

\section{Mutual Information (MI) Estimation}
\label{sec:background}
% introduce DIM and its theorem
In this section, we introduce how scalable estimation of mutual information is performed in terms of practical scenarios via mutual information neural estimation (MINE) \cite{belghazi2018mine} and the deep infomax (DIM) \cite{hjelm2018learning} described below.

The mutual information between two random variable $X$ and $Y$ is defined as:
\begin{equation}
    \label{eq:MI-def}
    \text{MI}(X, Y) = D_\text{KL}(p(X, Y) \parallel p(X)p(Y)),\nonumber
\end{equation}
where $D_{KL}$ is the Kullback-Leibler (KL) divergence between the joint distribution $p(X, Y)$ and the product of marginals $p(X)p(Y)$.

MINE estimates mutual information by training a classifier to distinguish between positive samples $(x, y)$ from the joint distribution and negative samples $(x, \bar{y})$ from the product of marginals.
Mutual information neural estimation (MINE) uses Donsker-Varadhan representation (DV) \cite{donsker1983asymptotic} as a lower-bound to estimate MI.
\begin{equation}
    \label{eq:DV}
    \text{MI}(X, Y) \geq \mathbb{E}_{\mathbb{P}}[g(x, y)] - \log( \mathbb{E}_{\mathbb{N}}[e^{g(x, \bar{y})}]),\nonumber
\end{equation}
where $\mathbb{E}_{\mathbb{P}}$ and $\mathbb{E}_{\mathbb{N}}$ denote the expectation over positive and negative samples respectively, and $g$ is the discriminator function that outputs a real number modeled by a neural network.

While the DV representation is the strong bound of mutual information shown in MINE, we are primarily interested in maximizing MI but not focusing on its precise value.
Thus DIM proposes an alternative estimation using Jensen-Shannon divergence (JS), which can be efficiently implemented using the cross-entropy (BCE) loss:
\begin{eqnarray}
    \label{eq:BCE}
    \text{MI}(X, Y)  &\geq& \mathbb{E}_{\mathbb{P}}[\log(g(x, y))] \\ &+& \mathbb{E}_{\mathbb{N}}[\log(1 - g(x, \bar{y}))].\nonumber
\end{eqnarray}
While two representations should behave similarly, considering that both act like classifiers with objectives maximizing the expected log-ratio of the joint over the product of marginals, it is found that the BCE loss empirically works better than the DV-based objective \cite{hjelm2018learning,ravanelli2018learning,velivckovic2018deep}. 
The reason may be that the BCE loss is bounded (i.e., its maximum is zero), making the convergence of the network more numerically stable.
In our experiments, we primarily use the JS representation to estimate mutual information.

Recently, \citet{tian2019contrastive} showed strong empirical performance through the improved multiview CPC training \cite{oord2018representation}, which shares many common ideas as mutual information maximization.
Inspired by their work, we modify (\ref{eq:BCE}) by first switching the role of $x$ and $y$ and summing them up:
\begin{eqnarray}
    \label{eq:MVBCE}
    \text{MI}(X, Y)  &\geq& \mathbb{E}_{\mathbb{P}}[\log(g(x, y))]  \\ 
    &+& \frac{1}{2} \mathbb{E}_{\mathbb{N}}[\log(1 - g(x, \bar{y}))] \nonumber\\
    &+& \frac{1}{2} \mathbb{E}_{\mathbb{N}}[\log(1 - g(\bar{x}, y))], \nonumber
\end{eqnarray}
where $(\bar{x}, y)$ is also the negative sample sampled from the product of marginals.

We empirically find that (\ref{eq:MVBCE}) gives the best performance, and more exploration about parameterization of MI is left as our future work.

\section{Methodology}
\label{sec:methodology}
In the extractive question answering dataset like SQuAD, the answer $A = \{a_1, \dots, a_M\}$ to the question $Q = \{q_1, \dots, q_K\}$ is guaranteed to be the span $\{p_{m}, \dots, p_{m+M}\}$ in the paragraph $P = \{p_{1}, \dots, p_{N}\}$.
Given $Q$ and $P$, the encoded representations from the QA system $M$ can be formulated as:
\begin{equation}
   \{r^q, r^p\} = \{r^q_1, \dots, r^q_K, r^p_1, \dots, r^p_N \} = M(Q, P),\nonumber
\end{equation}
where $r^q$ and $r^p$ are representations of the question and the passage respectively after the reasoning process in the QA system $M$. 

Most models then feed the passage representation $r^p$ to a single-layer neural network, obtain the span start and end probabilities for each passage word, and compute the loss $L_{span}$, which is the negative sum of log probabilities of the predicted distributions indexed by true start and end indices.

\begin{figure}[t!]
  \centering
  \includegraphics[width=\linewidth]{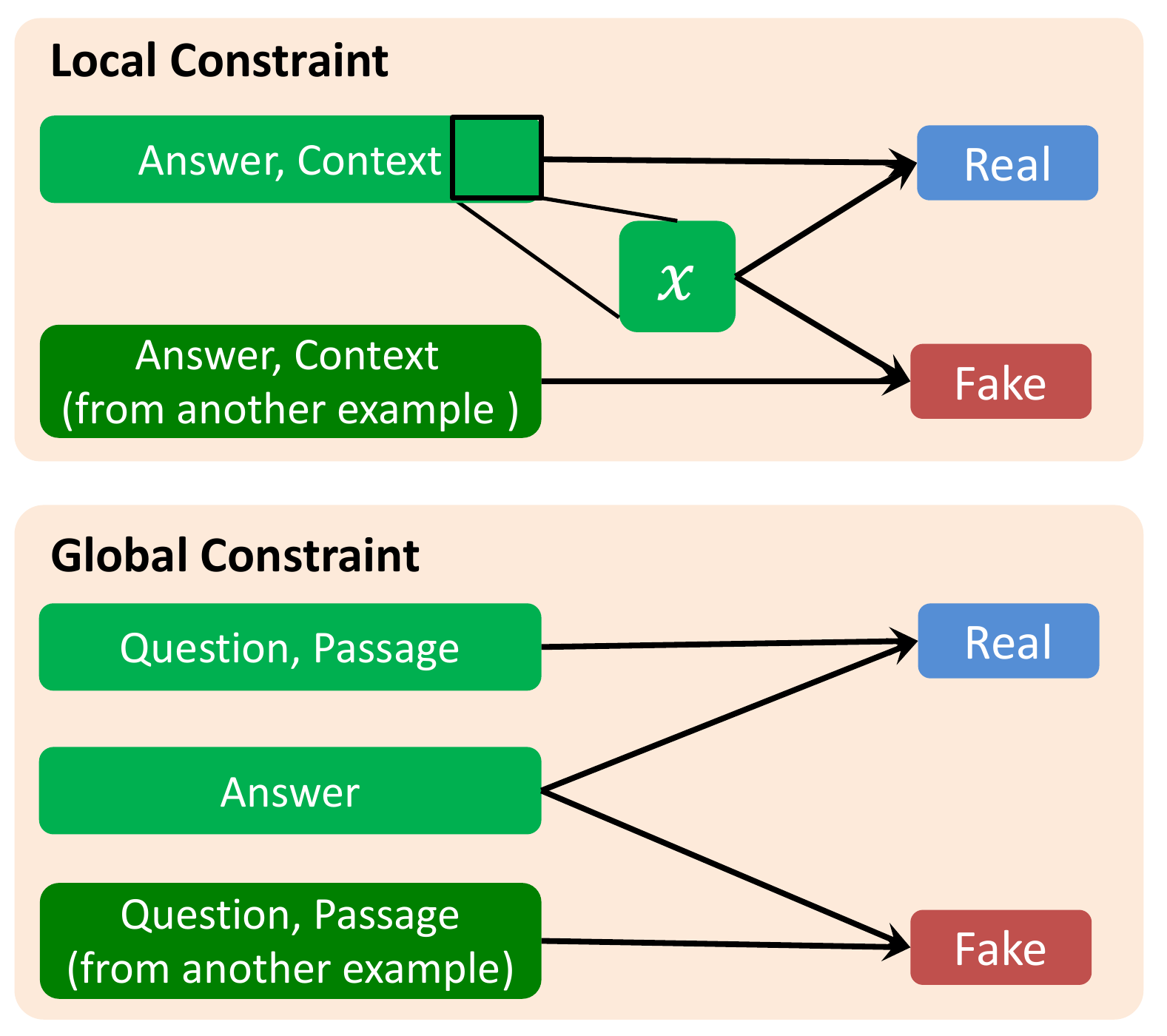}
  %\vspace{-1mm}
  \caption{Illustration of the LC and GC.}
  \label{fig:infomax}
%  \vspace{-5mm}
\end{figure}

Our QAInfomax aims at regularizing the QA system $M$ to not simply exploit the superficial biases in the dataset for answering questions.
Therefore, two constraints are introduced in order to guide the model learning.
\begin{enumerate}
    \item \textbf{Local Constraint (LC)}: each answer word representation $r^p_i$ in the answer representation $r^a = \{r^p_m, \dots, r^p_{m+M}\}$ should contain information about what the remaining answer words and its surrounding context are.
    \item \textbf{Global Constraint (GC)}: the summarized answer representation $s = S(r^a)$ should maximize the averaged mutual information to all other question representations in $r^q$ and passage representations in $r^p$, where $S$ is a summarization function described below. 
\end{enumerate}
%Intuitively, the model is expected to know \emph{what they have answered},
%and thus each generated answer word representation should have the information of each other.
Intuitively, the model is expected to \emph{choose the answer span after fully considering the entire question and paragraph}.
However, traditional QA models suffered the overstability problem, and tended to be fooled by distractor answers, such as the one containing an unrelated human name.
As \citet{lewis2018generative} argued, we also believe that the main reason is that QA models are only trained to predict start and end positions of answer spans.
Correlation in the dataset allows QA models to find shortcuts and ignore what the answer span looks like.
A learned behavior of traditional QA models can be viewed as a simple pattern matching, such as choosing the 5-length span after the word ``river'' if a question is about a river and the context talks about countries in European.

Following the intuition, two constraints \clocal and \cglobal are introduced to guide models to learn the desired behaviors.
%\clocal forces the model to generate answer span representation while maximizing the mutual information among words in a such span, and therefore model can no longer ignore what the words are in the chosen answer span.
%It is also natural for human that given part of the answer, they can more easily find what the remaining part is.
To prevent the model from only learning to match some specific word patterns to find the answer, \clocal forces the model to generate answer span representations while maximizing mutual information among words in the span and the context words surrounding the span.
By maximizing the mutual information between an answer word and \textit{all} of its context words, models need to incorporate the entire context into its decision process while choosing answers, and thus can be more robust to the adversarial sentences.
Then we further require models to maximize mutual information among answer words, so models can no longer ignore any word in the chosen answer span.

On the other hand, different from \clocal, which only focuses on the answer span and its context, \cglobal pushes the model to prefer answer representations carrying information that is globally shared across the whole input conditions $Q$ and $P$, because shortcuts do not necessarily appear near to the answer.
If the model only learns to leverage the correlation specific to the partial input, the MI of any input word without such relationship would \textit{not} increased.

The overview about two proposed constraints is illustrated in Figure~\ref{fig:infomax}.
The detail of two constraints and our QAInfomax regularizer is described below.

\subsection{Local Constraint}
\label{sec:local_consraint}

As shown in Section~\ref{sec:background}, the maximization of MI needs positive samples and negative samples drawn from joint distribution and the product of marginal distribution respectively.

In \clocal, because all answer word representations are expected to carry the information of each other and their contexts, we choose to maximize averaged MI between the sampled answer word representations and the whole answer sequence with its context words.
Specifically, a positive sample is obtained by pairing the sampled answer word representation $x \in r^a = \{r^p_m, \dots, r^p_{m+M}\}$ to all other answer and context words $r^c = \{r^p_{m-C}, \dots r^p_{m+M+C} \} \setminus \{x\}$, where $C$ is the hyperparameter defining how many context words for consideration.
Negative samples, on the other hand, are obtained by randomly sampling answer representation $\bar{r}^a = \{\bar{r}^p_{l}, \dots, \bar{r}^p_{l + L}\}$ and the corresponding $\bar{r}^c$ from other training examples.
Following (\ref{eq:MVBCE}), the objective for sampled $x$, $r^c$, $\bar{x} \in \bar{r}^a$ and $\bar{r}^c$ is formulated.
\begin{eqnarray}
    \label{eq:local}
        \text{LC}(x, r^c, \bar{x}, \bar{r}^c) &=& \frac{1}{|r^c|}\sum_{r^c_i \in r^c} \log(g(x, r^c_i)) \\
        &+& \frac{1}{2|\bar{r}^c|}\sum_{\bar{r}^c_j \in \bar{r}^c} \log(1 - g(x, \bar{r}^c_j))\nonumber \\
        &+& \frac{1}{2|r^c|}\sum_{r^c_i \in r^c} \log(1 - g(\bar{x}, r^c_i)).\nonumber
\end{eqnarray}

\begin{table*}[t!]
    \begin{center}
    %\small
    \begin{tabular}{|l|ccc|}
    \hline
        {\bf Model} & {\bf Original} & {\bf \textsc{AddSent}} & {\bf \textsc{AddOneSent}}\\
    \hline\hline
        BiDAF-S \cite{DBLP:journals/corr/SeoKFH16} & 75.5 & 34.3 & 45.7  \\
        %RaSOR \cite{DBLP:journals/corr/LeeKP016} & 81.1 & 39.5 & 49.5 \\
        %MPCM-S \cite{DBLP:journals/corr/WangMHF16} & 77.0 & 40.3 & 50.0 \\
        ReasoNet-S \cite{shen2017reasonet} & 78.2 & 39.4 & 50.3 \\
        Reinforced Mnemonic Reader-S \cite{hu2017reinforced} & 78.5 & 46.6 & 56.0 \\
        QANet-S \cite{yu2018qanet} & 83.8 & 45.2 & 55.7\\
        GQA-S \cite{lewis2018generative} & 83.7 & 47.3 & 57.8 \\
        FusionNet-E \cite{huang2017fusionnet} & 83.6 & 51.4 & 60.7  \\
        BERT-S \cite{devlin2018bert} & 88.5 & 51.0 & 63.4 \\
        \hline
        BERT-S + \model & \bf 88.6 & \bf 54.5 $^\dag$ & \bf 64.9 $^{\dag}$  \\
    \hline
    \end{tabular}
    \end{center}
%    \vspace{-3mm}
    \caption{\label{tab: main} F-measure on \textsc{AdversarialSquad} (S: single, E: ensemble). $^\dag$ indicates the significant improvement over baselines with p-value $< 0.05$.}
 %   \vspace{-5mm}
\end{table*}

\subsection{Global Constraint}
\label{sec:global_constraint}

Different from \clocal described above, \cglobal forces the learned answer representations $r^a$ to have information shared with all other question and passage representations. 
Here, we maximize the mutual information between the summarized answer vector $s = S(r^a)$ and $r_l \in r = \{r^q, r^p\} \setminus \{r^a\}$ pairs.
In the experiments, we use $S(r^a) = \sigma(\frac{1}{M}\sum r^a_i)$ as our summarization function, where $\sigma$ is the logistic sigmoid nonlinearity.

Specifically, a positive sample here is the pair of a answer summary vector $s = S(r^a)$ and all other word representations in $r$. 
%Negative samples are provided by pairing the summary $s$ with $\bar{r} = \{\bar{r}^q, \bar{r}^p\} \setminus \{\bar{r}^a\}$, where $\bar{r}^q$ and $\bar{r}^p$ are question and passage representations from an alternative training example respectively.
Negative samples are provided by sampling question, passage and answer representations $\{\bar{r}^q, \bar{r}^p, \bar{r}^a\}$ from an alternative training example. 
Then we pair the summary $s$ with $\bar{r} = \{\bar{r}^q, \bar{r}^p\} \setminus \{\bar{r}^a\}$, and $\bar{s} = S(\bar{r}^a)$ with $r$.

Similar to (\ref{eq:local}), the objective for the sampled $s$, $r$, $\bar{s}$ and $\bar{r}$ is:
\begin{eqnarray}
    \label{eq:global}
        \text{GC}(s, r, \bar{s}, \bar{r}) &=& 
        \frac{1}{|r|}\sum_{r_i \in r} \log(g(s, r_i))  \\ 
        &+& \frac{1}{2|\bar{r}|}\sum_{\bar{r}_j \in \bar{r}} \log((1 - g(s, \bar{r}_j))) \nonumber\\
        &+& \frac{1}{2|r|}\sum_{r_i \in r} \log((1 - g(\bar{s}, r_i))).\nonumber
\end{eqnarray}

\subsection{\model}
\label{sec:qainfomax}
In our proposed model, we combine two objectives and formulate the model as the complete \model regularizer.
For each training batch consisting of training examples $\{\{Q_1, P_1, A_1\}, \dots \{Q_B, P_B, A_B\} \}$, we pass the batch into the model $M$ and obtain representations $\{\{r^{q}_{1}, r^{p}_{1}, r^{a}_{1}\}, \dots, \{r^{q}_{B}, r^{p}_{B}, r^{a}_{B}\}\}$. 
Note that we abuse the subscripts to denote the example index in the batch for simplicity.

Then we shuffle the whole batch to obtain negative examples $\{\{\bar{r}^{q}_{1}, \bar{r}^{p}_{1}, \bar{r}^{a}_{1}\}, \dots, \{\bar{r}^{q}_{B}, \bar{r}^{p}_{B}, \bar{r}^{a}_{B}\}\}$. The complete objective $L_{info}$ for \model becomes: 
\begin{equation}
    \label{eq:qainfomax}
    -\frac{1}{B} \sum_{i=1}^{B} (\alpha LC(x_i, r^c_i, \bar{x}_i, \bar{r}^c_i) + \beta GC(s_i, r_i, \bar{r}_i)),\nonumber
\end{equation}
where $x_i$ and $\bar{x}_i$ are the representation sampled from $r^a_i$ and $\bar{r}^a_i$, $r^c_i$ and $\bar{r}^c_i$ are $r^a_i$ and $\bar{r}^a_i$ expanded with its context words respectively, $s_i$ and $\bar{s}_i$ are the summary vectors of $r^a_i$ and $\bar{r}^a_i$, and $\alpha$ and $\beta$ are hyperparameters.

Combined with \model as a regularizer, the final objective of the model becomes
\begin{equation}
    L = L_{span} + \gamma L_{info},
\end{equation}
where $L_{span}$ is the answer span prediction loss and $\gamma$ is the regularize strength.
The objective can be optimized through the simple gradient descent.

\section{Experiments}
\label{sec:exp}

To evaluate the effectiveness of the proposed QAInfomax, we conduct the experiments on a challenging dataset, Adversarial-SQuAD.

\subsection{Setup}
\label{sec:setup}

BERT-base \cite{devlin2018bert} is employed as our QA system $M$ in the experiments, where we set the same hyperparameters as one released in SQuAD training\footnote{We use PyTorch \cite{paszke2017automatic} reimplementation for experiments: \url{https://github.com/huggingface/pytorch-pretrained-BERT}.}.
%As BERT is a powerful pretrained language understanding model, the fine-tune process can be done in few hours, which is suitable for our preliminary experiments. 

We set $C$, $\alpha$, $\beta$ and $\gamma$ to be $5$, $1$, $0.5$, $0.3$ respectively in all experiments, and add the proposed QAInfomax into the BERT model as described above.
The discriminator function $g$ is the bilinear function similar to the scoring used by \citet{oord2018representation}:
\begin{equation}
    \label{eq:bilinear}
    g(x, y) = x^TWy,
\end{equation}
where $W$ is a learnable scoring matrix.

We train the BERT model with the proposed \model on the orignal SQuAD dataset, and use Adversarial-SQuAD to test the robustness of the augmented model.
Only \textsc{AddSent} and \textsc{AddOneSent} metrics are reported for the comparison with previous models, because most previous models did not report their \textsc{AddAny} and \textsc{AddCommon} scores.
Briefly, for each example, \textsc{AddSent} runs the model $M$ on every human-approved adversarial sentence, picks the one that makes the model give the worst answer and returns that score.
\textsc{AddOneSent}, on the other hand, only picks a random human-approved adversarial sentence. 
The numbers reported in all experiments are the best number across at least three runs.

\begin{table}[t!]
    \begin{center}
%    \small
    \begin{tabular}{|l|cc|}
    \hline
        {\bf Model} & {\bf Adversary F1} & {\bf Speed (iter/s)} \\
    \hline\hline
        BERT & 51.0 / 63.4 & 3.80 \\
        + LC & 53.6 / 64.2 & 3.51 \\
        + GC & 52.2 / 63.7 & 2.75\\
        + LC + GC & \bf 54.5 / \bf 64.9 & 2.72 \\
    \hline
    \end{tabular}
    \end{center}
%    \vspace{-3mm}
    \caption{\label{tab: ablation study} Ablation study with F1 scores  on \textsc{AddSent} / \textsc{AddOneSent}. The speed is measured on RTX 2080Ti.}
 %   \vspace{-2mm}
\end{table}

\subsection{Results}
\label{sec:results}

Table \ref{tab: main} reports model performance on Adversarial-SQuAD. %compared to other question answering systems.
It can be found that \model yields substantial improvement over the vanilla BERT model, and achieves the state-of-the-art performance on both \textsc{AddSent} and \textsc{AddOneSent} metrics.\footnote{Note that \citeauthor{wang2018robust} modified distractor paragraphs and added them into training data, so we do not compare with them, because we only use the original SQuAD training data.}
\model obtains larger improvement on the \textsc{AddSent}, which picks the worst scores of the model.
It shows the effectiveness of our \model in terms of forcing the model to ignore simple correlation in the data and learn the more human-like reasoning processes.
It is worth to note that while \model mitigates the overstability problem and improves the robustness to adversarial examples, it does not hurt the original performance of the QA system, demonstrating the benefit for the practical usage.
Some example results from the Adversarial-SQuAD dataset can be found in the Appendix, where adversarial distracting sentences are shown in italic blue fonts.

Table \ref{tab: ablation study} shows the ablation study of our proposed QAInfomax, where two proposed constraints are both important for achieving such results. 
We also show the training speed of the proposed method and its limitation, where the \cglobal objective degrades the training speed by $28\%$.
The reason is that \cglobal measures the averaged MI over the \emph{whole} question and passage representations, which may include a long sequence of vectors.

% add different sampling strategy? training speed compare?
\begin{table}[t!]
    \begin{center}
%    \small
    \begin{tabular}{|l|cc|}
    \hline
        {\bf Function} & {\bf \textsc{AddSent}} & {\bf \textsc{AddOneSent}}  \\
    \hline\hline
        Mean & \bf 52.2 & \bf 63.7\\
        Max & 52.0 & 63.3 \\
        Sample & \bf 52.2 & 63.0 \\
    \hline
    \end{tabular}
    \end{center}
%    \vspace{-3mm}
    \caption{\label{tab: variants} Different summarization functions for \cglobal.}
%    \vspace{-5mm}
\end{table}

Considering that the summarization function $S$ plays an important role in GC, we explore its different variants in Table \ref{tab: variants}:
\begin{itemize}
    \item Mean: $\sigma(\frac{1}{M}\sum r^a_i)$
    \item Max: $\sigma(maxpool(r^a))$
    \item Sample: randomly sample one $r^a_i \in r^a$ %as the summary
\end{itemize}
According to the experimental results, Mean performs the best while Max and Sample has the competitive performance, showing the great robustness of the proposed methods to different architecture choices.

\section{Conclusion}
\label{sec:conclusion}
This paper presents a novel regularizer based on MI maximization for question answering systems named QAInfomax, which helps models be not stuck with superficial correlation in the data and improves its robustness.
The proposed QAInfomax is flexible to apply to different machine comprehension models.
The experiments on Adversirial-SQuAD demonstrate the effectiveness of our model, and the augmented model achieves the state-of-the-art results.
In the future, we will investigate more methods for reducing the limitations of QAInfomax and improving the capability of generalization in QA systems.
%Considering that QAInfomax still has many limitations such as the degradation of training speed, in the future we will investigate more methods to improve the generalization ability of QA systems.

\section*{Acknowledgements}
We would like to thank reviewers for their
insightful comments on the paper.
This work was financially supported from the Young Scholar Fellowship Program by Ministry of Science and Technology (MOST) in Taiwan, under Grant 108-2636-E-002-003.

\bibliography{emnlp-ijcnlp-2019}
\bibliographystyle{acl_natbib}

\clearpage
\appendix

%\section{Qualitative Analysis}
%Here we present some examples from the Adversarial-SQuAD dataset, and adversarial distracting sentences are shown in \textcolor{blue}{blue}.

\begin{figure*}[t!]
%\small
\begin{framed}
%\footnotesize
\textbf{Article:} Force

\textbf{Paragraph:}
     A static equilibrium between two forces is the most usual way of measuring forces, using simple devices such as weighing scales and spring balances. For example, an object suspended on a vertical spring scale experiences the force of gravity acting on the object balanced by a force applied by the "spring reaction force", which equals the object's weight. Using such tools, some quantitative force laws were discovered: that the force of gravity is proportional to volume for objects of constant density (widely exploited for millennia to define standard weights); Archimedes' principle for buoyancy; Archimedes' analysis of the lever; Boyle's law for gas pressure; and Hooke's law for springs. These were all formulated and experimentally verified before Isaac Newton expounded his Three Laws of Motion.
     \textcolor{blue}{\it Jeff Dean expounded on the Four Regulations of Action.}

     \textbf{Question:} Who expounded the Three Laws of Motion?  
     
     \textbf{Ground Truth:} \textcolor{teal} {Isaac Newton}
     
     \textbf{BERT Original Prediction:} \textcolor{teal} {Isaac Newton}
     
     \textbf{BERT Prediction under adversary:} \textcolor{red}{Jeff Dean}
     
     \textbf{BERT + QAInfomax Prediction:} \textcolor{teal}{Issac Newton}
\end{framed}
\begin{framed}
%\footnotesize
\textbf{Article:} Islamism  

\textbf{Paragraph:}
     The views of Ali Shariati, ideologue of the Iranian Revolution, had resemblance with Mohammad Iqbal, ideological father of the State of Pakistan, but Khomeini's beliefs is perceived to be  placed somewhere between beliefs of Sunni Islamic thinkers like Mawdudi and Qutb. He believed that complete imitation of the Prophet Mohammad and his successors such as Ali for restoration of Sharia law was essential to Islam, that many secular, Westernizing Muslims were actually agents of the West serving Western interests, and that the acts such as "plundering" of Muslim lands was part of a long-term conspiracy against Islam by the Western governments.
     \textcolor{blue}{\it The short term agenda of hamster was the acts of plundering Islamic lands by the East.}

     \textbf{Question:} Who expounded the Three Laws of Motion?  
     
     \textbf{Ground Truth:} \textcolor{teal} {conspiracy}
     
     \textbf{BERT Original Prediction:} \textcolor{teal} {conspiracy against Islam}
     
     \textbf{BERT Prediction under adversary:} \textcolor{red}{The short term agenda of hamster }
     
     \textbf{BERT + QAInfomax Prediction:} \textcolor{teal}{conspiracy against Islam}
\end{framed}
\begin{framed}
%\footnotesize
\textbf{Article:} Force

\textbf{Paragraph:}
     Newton's First Law of Motion states that objects continue to move in a state of constant velocity unless acted upon by an external net force or resultant force. This law is an extension of Galileo's insight that constant velocity was associated with a lack of net force (see a more detailed description of this below). Newton proposed that every object with mass has an innate inertia that functions as the fundamental equilibrium "natural state" in place of the Aristotelian idea of the "natural state of rest". That is, the first law contradicts the intuitive Aristotelian belief that a net force is required to keep an object moving with constant velocity. By making rest physically indistinguishable from non-zero constant velocity, Newton's First Law directly connects inertia with the concept of relative velocities. Specifically, in systems where objects are moving with different velocities, it is impossible to determine which object is ``in motion'' and which object is ``at rest''. In other words, to phrase matters more technically, the laws of physics are the same in every inertial frame of reference, that is, in all frames related by a Galilean transformation
     \textcolor{blue}{\it The Rosetta laws of physics refer to an object in motion and rest.}

     \textbf{Question:} What are the laws of physics of Galileo, in reference to objest in motion and rest? 
        
     \textbf{Ground Truth:} \textcolor{teal} {the laws of the physics are the same in every inertial frame of reference, that is, in all frames related by a Galilean transformation.}
     
     \textbf{BERT Original Prediction:} \textcolor{teal} {the same in every inertial frame of reference, that is, in all frames related by a Galilean transformation.}
     
     \textbf{BERT Prediction under adversary:} \textcolor{red}{Rosetta laws}
     
     \textbf{BERT + QAInfomax Prediction:} \textcolor{red}{Rosetta laws}
\end{framed}
%\caption{An example from the SQuAD dataset. BERT originally gets the answer correct, but is fooled by adversarial distractiing sentence \textcolor{blue}{(in blue)}.}
%\label{fig:exp}
\end{figure*}

\end{document}